# E2E-AFG: An End-to-End Model with Adaptive Filtering for Retrieval-Augmented Generation

Yun Jiang, Zilong Xie, Wei Zhang, Yun Fang and Shuai Pan[*]

Advanced Institute of Information Technology, Peking University, China
pans@aiit.org.cn

**Abstract.** Retrieval-augmented generation methods often neglect the quality of content retrieved from external knowledge bases, resulting in irrelevant information or potential misinformation that negatively affects the generation results of large language models. In this paper, we propose an end-to-end model with adaptive filtering for retrieval-augmented generation (E2E-AFG), which integrates answer existence judgment and text generation into a single end-to-end framework. This enables the model to focus more effectively on relevant content while reducing the influence of irrelevant information and generating accurate answers. We evaluate E2E-AFG on six representative knowledge-intensive language datasets, and the results show that it consistently outperforms baseline models across all tasks, demonstrating the effectiveness and robustness of the proposed approach. [1]

## 1 Introduction

The remarkable natural language understanding and generation capabilities demonstrated by Large Language Models (LLMs) have led to their success in knowledge-intensive tasks, such as open-domain question answering and fact verification [4, 28, 1]. However, LLMs are prone to generating hallucinatory content that contains factual errors in the absence of supporting documentation. To address this issue, [21] proposed the retrieval-augmented generation (RAG) method, which involves retrieves relevant context from external knowledge bases to provide additional evidence for LLMs when answering input queries. Other approaches [31] directly utilize a pre-trained LLM to generate a relatively accurate pseudo-answer as an extended document for the input query. However, these methods often fail to adequately consider the quality of the retrieved or generated content, which may include distracting irrelevant content or erroneous information, leading LLMs to still produce hallucinatory answers.

Earlier studies [32, 23] attempted to select more relevant content by re-ranking the retrieved contexts, but they may still contain irrelevant information. [7] achieved automatic decontextualization of sentences through training a coreference resolution model, although this requires extensive manual annotation efforts. Recent research, such as HyDE [12], employs unsupervised contrastive learning where an encoder's dense bottleneck acts as a lossy compressor to filter out hallucinatory content. FILCO [33] trains a filtering model to remove irrelevant contexts, improving the quality of the context provided to the generation model. However, these methods typically involve multiple independent models and complex preprocessing operations, which not only increase system complexity but also elevate training and inference costs.

---

[*] Corresponding Author
[1] Our code is available at: https://github.com/XieZilongAI/E2E-AFG

To address the aforementioned issues, we propose an End-to-End Model with Adaptive Filtering for Retrieval-Augmented Generation (E2E-AFG), which integrates classification and generation tasks into an end-to-end framework, allowing the model to simultaneously learn context filtering and answer generation. Specifically, we first employ a pre-trained large language model to generate a pseudo-answer related to the input query, enriching the content. We then apply three context filtering strategies to obtain silver classification labels. The construction of the end-to-end model is based on the generation model, augmented with a classification module that employs a cross-attention mechanism to predict whether sentences in the context contain answers, enabling the model to answer the input query based on a certain judgment of the context.

We conducted experiments on six knowledge-intensive language datasets, covering three tasks: question answering (Natural Questions [19], TriviaQA [17], HotpotQA [36], ELI5 [10]), fact verification (FEVER [30]), and knowledge-based dialogue generation (Wizard of Wikipedia [9]). Compared to baseline models, our approach achieved state-of-the-art results across all six datasets, with improvements ranging from +0.13 to +1.83 points, validating the effectiveness of the proposed method.

## 2  Related Work

**Retrieval-Augmented Generation.** Early research methods such as REALM [13] and RAG [21], laid the foundation for the field of retrieval-augmented generation (RAG) by combining retrievers with large language models (LLMs). Subsequently, RETRO [3] introduced the concept of training language models on fixed retrievers, while Atlas [16] further explored dedicated loss functions and training strategies, achieving improved results, particularly in few-shot learning scenarios. Recent studies have shifted towards optimizing the retrieval component while leveraging pre-trained, fixed LLMs. For instance, RePlug [34] and In-context RALM [29] demonstrated that fine-tuning the retrieval module can surpass end-to-end trained models in certain tasks, such as question answering. In contrast, SAIL [22] integrated real search engines with information denoising processes, aiming to enhance the relevance and accuracy of retrieval results, showcasing potential in broader application contexts. Our work seeks to enhance attention to reliable information by performing answer existence judgment on the retrieved passages prior to generation, thereby reducing the interference caused by irrelevant information.

**Retrieval Content Filtering Strategies.** In knowledge-intensive tasks, post-processing of retrieved content is crucial for enhancing system performance, with common practices including re-ranking and context filtering. In early studies, [32] and [20] explored passage re-ranking methods based on BiLSTM, while [23] and [27] employed BERT-based cross-encoders to achieve more precise passage re-ranking. Subsequently, [26] proposed a method for re-ranking passages by updating the query, and [15] directly applied heuristic re-ranking to the answers. In recent years, several context filtering strategies have been introduced. For example, FILCO [33] trains a context filtering model to perform fine-grained sentence-level filtering on the retrieved passages. Multi-Meta-RAG [25] utilizes a specific set of domain queries and formats to select the most relevant documents through database filtering. In contrast, our approach constructs a single end-to-end model that can simultaneously perform context filtering and answer generation.

**Multi-task Learning.** Multi-task learning (MTL) enhances overall model performance by jointly learning multiple tasks, allowing it to capture the correlations and shared features among tasks [5]. In natural language processing applications, MTL not only leverages task relevance to mitigate issues of data scarcity and model overfitting but also improves the generalization capability of the model. For instance, [6] proposed a hierarchical multi-task learning approach that enhances the model's ability to capture inter-task dependencies. ROM [11] introduced a generalizable Retrieval Optimized Multi-task framework that reduces the model's parameters. Our method applies MTL to the retrieval-augmented generation domain by jointly learning binary classification and generation tasks, enabling the model to acquire context filtering and answer generation capabilities.

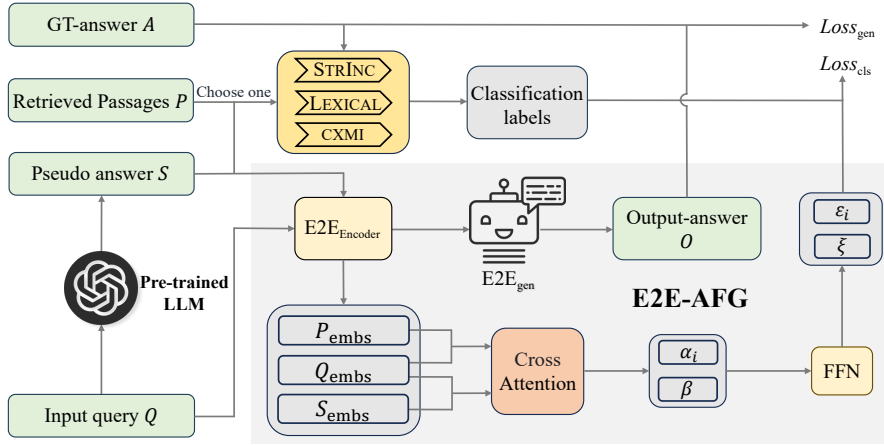

**Fig. 1:** The overall architecture diagram of the proposed method.

## 3 Method

**Problem Statement.** In knowledge-intensive tasks, each entry consists an input query $Q$, a ground truth answer $A$, and a set of retrieved passages $P = \{p_i\}_{i=1}^{K}$ from a database. We provide the generator with one or more passages along with a pre-generated pseudo-answer $S$ to generate a response to the query $Q$. Specifically, in the question-answering tasks, $Q$ and $A$ are natural language questions and their corresponding ground truth answers; in the fact verification tasks, $Q$ is a statement and $A \in$ {SUPPORTS, REFUTES} indicates the correctness of the statement; in the knowledge-based dialogue generation tasks, $Q$ consists of a dialogue history, and $A$ is a response that accurately continues the conversation.

**Overview.** The overall architecture of our proposed method is illustrated in Fig. 1. First, a pre-trained large language model generates a pseudo-answer $S$ for the query $Q$. Next, the query $Q$, the retrieved set of passages $P$, and the pseudo-answer $S$ are input into the E2E-AFG model, where both generation and binary classification tasks are performed. The generation task utilizes the generator $E2E_{gen}$ to produce an answer. The binary classification task employs $E2E_{Encoder}$ to obtain embeddings for the three inputs, which are then processed through cross-attention and a feedforward neural network to predict the category scores. Finally, the cross-entropy loss for both the generation and binary classification tasks is computed. This approach allows for the update of the internal parameters of the shared $E2E_{Encoder}$, implicitly learning a filtering capability that prioritizes sentences more likely to contain answers while reducing interference from irrelevant sentences.

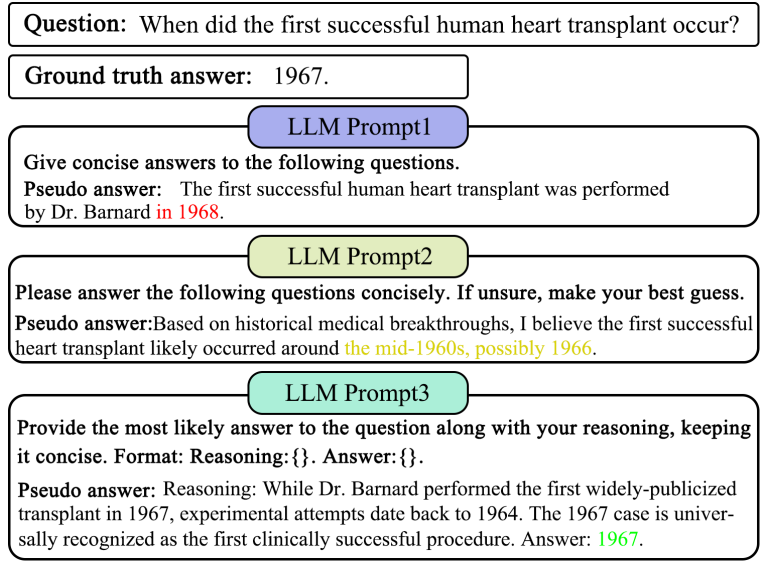

**Fig. 2:** Three kinds of LLM prompts and their generated pseudo-answer examples.

### 3.1 Generating Pseudo-Answers

In retrieval-augmented generation frameworks, semantic discrepancies between retrieved results and original queries may lead to model misinterpretation of knowledge boundaries. To enhance the model's capability in understanding and filtering out-of-distribution contextual information, we employ pretrained language models to generate pseudo-answers with semantic variation features, thereby constructing diversified training samples. As illustrated in Fig. 2, we propose a progressive prompting strategy comprising three types of prompts: 1) direct generation of concise answers potentially containing semantic deviations; 2) reasoned speculation under incomplete information conditions; 3) simultaneous output of conclusions and their logical derivations. This multi-tiered mechanism simulates heterogeneous contextual qualities in real retrieval scenarios. Through controlled noisy data generation with this approach, we establish robust evidence evaluation during training, thereby improving error detection in retrieved results.

### 3.2    Obtaining Silver Classification Labels

To determine whether the retrieved passage set $P$ and the generated pseudo-answer $S$ contain answers, we introduce three context filtering methods based on [33]: (i) String Inclusion (STRINC): checking if the context directly contains the ground truth answer; (ii) Lexical Overlap (LEXICAL): measuring the overlap of words between the context and the ground truth answer; and (iii) Conditional Cross-Mutual Information (CXMI): assessing the likelihood of the generator producing the ground truth answer given the context. For a specific task, we select the most appropriate filtering method to obtain silver classification labels. For instance, in question-answering tasks, we may use StrInc to evaluate whether each passage or pseudo-answer contains the ground truth answer. In contrast, for fact extraction tasks, where the ground truth answer resembles a boolean value and cannot be assessed using the first two methods, we employ CXMI to compute the corresponding probability and set a threshold $t_0$ to derive the silver classification label. We concatenate the obtained labels with the ground truth answer $A$ to facilitate loss calculation.

### 3.3    Generation Task

For each training sample $(Q, A, P, S)$, we first insert a special character between the different fields to ensure they can be distinguished after encoding with $\text{E2E}_{\text{Encoder}}$. We then input the encoded query $Q_{\text{embs}}$, the retrieved passage set $P_{\text{embs}}$, and the pseudo-answer $S_{\text{embs}}$ into $\text{E2E}_{\text{gen}}$ to produce the output answer $O$. The sequence probability is calculated as follows:

$$P_o(O|Q, P, S) = \prod_{i=1}^{L} p(o_i|O_{<i}, Q, P, S) \qquad (1)$$

where $o_i$ represents the $i$-th token of the generated output $O$, and $L$ is the final output length. To simplify the notation, we continue to use $Q, P, S$ in place of $Q_{\text{embs}}, P_{\text{embs}}$, and $S_{\text{embs}}$ respectively in the equations above and in the subsequent content. The loss function for the generation task is calculated as follows:

$$L_{\text{gen}} = -\sum_{i=1}^{L} \log p(o_i^{gt}|O_{<i}, Q, P, S) \qquad (2)$$

where $o_i^{gt}$ denotes the $i$-th token of the ground truth answer $A$.

### 3.4 Classification Task

To enhance the model's context filtering capability, we introduce a classification module specifically designed to determine whether the input context contains the answer. The generator and the classification module share the same encoder E2E$_{\text{Encoder}}$, allowing the classification model to indirectly improve the model's context filtering capabilities by influencing the encoder's parameters.

The classification module comprises two main components: cross-attention layer, and feedforward neural network. First, the encoded query $Q$, each retrieved passage $p_i$, and the pseudo-answer $S$ are fed into the cross-attention layer. In this layer, the model computes the attention weights between $Q$ and $p_i$, as well as between $Q$ and $S$, generating cross-attention representations:

$$\alpha_i = \text{softmax}\left(\frac{Qp_i^T}{\sqrt{d_k}}\right) p_i \tag{3}$$

$$\beta = \text{softmax}\left(\frac{QS^T}{\sqrt{d_k}}\right) S \tag{4}$$

where $d_k$ is the dimensionality of the encoder's feature channels.

Next, the generated cross-attention representations are fed into a feedforward neural network to predict two binary classification results:

$$\varepsilon_i = \text{FFN}(\alpha_i), \quad \xi = \text{FFN}(\beta) \tag{5}$$

where FFN denotes a two-layer feedforward neural network. The loss function for the classification task is defined as the cross-entropy:

$$L_{\text{cls}} = \sum_{i=1}^{K} -(\log \varepsilon_i^{gt}) + \log \xi^{gt} \tag{6}$$

Here, $\varepsilon_i^{gt}$ and $\xi^{gt}$ represent the predicted probability values corresponding to the ground truth classes of each passage $p_i$ and the pseudo-answer $S$, respectively, while $K$ is the number of retrieved passages.

### 3.5 Model Training

During the training process, we simultaneously optimize the loss functions of both the generator and the classification module. The overall loss function is defined as a weighted sum of the two losses:

$$L_{\text{TOTAL}} = (1 - \sigma)L_{\text{gen}} + \sigma L_{\text{cls}} \tag{7}$$

where $L_{\text{gen}}$ is the loss from the generator, $L_{\text{cls}}$ is the loss from the classification module, and $\sigma$ is the weighting factor.

To further enhance the training efficiency and performance of the model, we employ Low-Rank Adaptation (LoRA) [14] techniques, which add low-rank matrices to the weight matrices of the pre-trained model for fine-tuning. This approach reduces computational overhead and accelerates the training process.

## 4    Experiments

### 4.1    Datasets and Evaluation Metrics

As shown in Table 1, we evaluate six retrieval-augmented knowledge-intensive language datasets constructed from Wikipedia articles as supporting evidence. Each dataset is partitioned into training (train), development (dev), and test sets. Exact Match (EM), which quantifies the percentage of predictions identical to the ground-truth answers; Unigram $F_1$ ($F_1$), computing the harmonic mean of precision and recall through word-level overlap between predictions and references; Accuracy (Acc), reflecting the ratio of correct predictions to total predictions; and Top-20 recall [2], which verifies whether the answer string exists within the top-20 retrieved passages (for Natural Questions [19] and TriviaQA-unfiltered [17]) or originates from annotated source articles in the KILT benchmark [24] (for FEVER [30] and Wizard of Wikipedia [9]).

Table 1: Statistics and evaluation metric for six datasets.

| Dataset | # Examples | | | Evaluation metric | Top-20 recall (%) |
|---|---|---|---|---|---|
| | train | dev | test | | |
| Natural Questions | 79,168 | 8,757 | 3,610 | EM | 82.1 |
| TriviaQA-unfiltered | 78,785 | 8,837 | 11,313 | EM | 75.2 |
| FEVER | 104,966 | 10,444 | 10,100 | Acc | 98.1 |
| HotpotQA | 88,924 | 5,947 | 5,631 | $F_1$ | 63.5 |
| ELI5 | 273,036 | 3,098 | 2,367 | $F_1$ | 56.5 |
| Wizard of Wikipedia | 63,734 | 3,054 | 2,944 | $F_1$ | 96.2 |

Open-Domain Question Answering employs the Natural Questions (NQ) and TriviaQA-unfiltered (TQA) datasets, comprising Wikipedia-derived questions paired with answers truncated to five tokens. Fact Verification utilizes the FEVER dataset, where claims are labeled as "SUPPORTS" or "REFUTES" based on alignment with Wikipedia evidence. Multi-Hop Question Answering leverages HotpotQA, featuring 113K complex queries requiring cross-passage reasoning. Long-Form Question Answering involves ELI5, containing 270K open-ended Reddit queries demanding multi-sentence explanations. Lastly, Knowledge-Based Dialogue Generation uses the Wizard of Wikipedia (WoW) dataset to generate responses grounded in dialogue history and Wikipedia-sourced knowledge.

### 4.2    Implementation Details

We loaded the model checkpoints from HuggingFace Transformers [35], using FLAN-T5-xl [8] as our backbone model architecture. Pseudo-answers are generated using the Llama-3 model with a mixture of three prompt types, with generation length limited to 200 tokens. With our primary focus on post-processing operations for retrieved content, we pre-process each query in the dataset by extracting the top-5 most relevant paragraphs from Wikipedia using an adversarial Dense Passage Retriever (DPR) [18]. To obtain silver classification labels, we adopted the optimized settings from FILCO, using

STRINC for NQ and TQA, LEXICAL for WoW, and CXMI for FEVER, HotpotQA, and ELI5, with a threshold $t_0$ set to 0.5.

For the generator E2E$_{gen}$, we allowed a maximum input sequence length of 512 tokens during both training and inference. We generated up to 64 tokens for open-domain question answering, multi-hop question answering, fact verification, and dialogue generation tasks, and up to 256 tokens for long-form question answering. We used greedy decoding to produce the final answers. Regarding model parameters, we set the encoder's feature channel dimension $d_k$ to 2048, trained for 3 epochs, with a learning rate of 5e−5 and a batch size of 8. The weight factor $\sigma$ was set to 0.2.

### 4.3  Baseline Methods

In this section, we introduce three baseline methods: FULL [21], HyDE [12], and FILCO [33], along with the proposed E2E-AFG and SILVER configurations. To ensure a fair comparison, we employed the same backbone model architecture across all methods as that used in our proposed E2E-AFG.

FULL: A common approach in retrieval-augmented generation where all passages, including pseudo-answers, are input into the generation model with the query.

HyDE: Filters passages through a dense bottleneck using unsupervised contrastive learning, encoding them before inputting into the generation model.

FILCO: Uses a trained model to filter sentences within passages, passing only the selected sentences to the generation model.

E2E-AFG: Ours end-to-end model potentially assesses the existence of answers for the input passages before feeding all passages into the model for answer generation.

SILVER: This configuration inputs only those passages labeled as containing an answer, testing the performance upper bound of E2E-AFG.

Table 2: Comparison with baseline methods using top-1 retrieved passages.

| Method  | NQ    | TQA   | FEVER | HotpotQA | ELI5  | WoW   |
|---------|-------|-------|-------|----------|-------|-------|
| FULL    | 41.64 | 60.90 | 88.32 | 59.58    | 67.50 | 65.73 |
| HyDE    | 43.37 | 62.28 | 90.27 | 60.62    | 71.38 | 67.60 |
| FILCO   | 46.65 | 64.33 | 94.46 | 62.71    | 74.99 | 70.12 |
| E2E-AFG | **48.48** | **65.99** | **95.45** | **64.39** | **75.12** | **71.47** |
| SILVER  | 51.77 | 68.73 | 96.64 | 65.50    | 77.89 | 72.68 |

### 4.4  Comparison with Baseline Methods

Table 2 presents the experimental results of E2E-AFG across six datasets, demonstrating that our model outperforms the baseline models in all cases. Specifically, for extractive question-answering tasks NQ and TQA, we achieved improvements of at least 1.83% and 1.56% in EM, respectively. This indicates that our model focuses more on credible passages and reduces attention to irrelevant information, thereby generating more accurate answers. In the fact verification task FEVER, we attained an accuracy increase of at least 1.09%. For the complex multi-hop question-answering task HotpotQA and the long-form question-answering task ELI5, we observed improvements of at least 1.68% and 0.13% in $F_1$ score, respectively. We hypothesize that the relatively modest performance gain on ELI5 may be due to the fact that it requires detailed, lengthy answers, while the generated pseudo-answers tend to be relatively brief, limiting the model's filtering capabilities. Additionally, in the dialogue generation task WoW, we improve the $F_1$ score by at least 1.35%. Furthermore, the performance of E2E-AFG approaches the upper bound performance of SILVER, indicating its exceptional capabilities in context filtering and text generation, allowing it to achieve near-optimal results without relying on specific annotations.

Table 3: The impact of different modules on the overall performance of E2E-AFG.

| Method | NQ | FEVER | WoW |
|---|---|---|---|
| Metric | EM | Acc | $F_1$ |
| Ours | **48.48** | **95.45** | **71.47** |
| - pseudo answer | 44.76 | 92.63 | 68.35 |
| - cross attention layer | 43.60 | 91.02 | 67.81 |
| - classification module | 40.03 | 87.52 | 65.12 |

Table 4: The recall rates of pseudo-answers generated by different prompts.

| Dataset | Recall (%) | | |
|---|---|---|---|
| | Prompt1 | Prompt2 | Prompt3 |
| Natural Questions | 40.3 | 45.6 | **46.8** |
| TriviaQA-unfiltered | 51.0 | **57.4** | 57.2 |
| FEVER | 62.8 | 63.7 | **65.3** |
| HotpotQA | 12.5 | 15.6 | **16.6** |
| ELI5 | 9.3 | 11.9 | **13.4** |
| Wizard of Wikipedia | 28.7 | 30.2 | **30.5** |

Table 5: The impact of different top-K retrieved passages on the generated results.

| Method | NQ | | | FEVER | | | WoW | | |
|---|---|---|---|---|---|---|---|---|---|
| | top-1 | top-3 | top-5 | top-1 | top-3 | top-5 | top-1 | top-3 | top-5 |
| FULL | 41.64 | 50.84 | 52.22 | 88.32 | 88.26 | 87.34 | 65.73 | 65.86 | 64.34 |
| HyDE | 43.37 | 52.91 | 58.77 | 90.27 | 91.69 | 91.82 | 67.60 | 68.07 | 68.15 |
| FILCO | 46.65 | 54.38 | 62.03 | 94.46 | 93.83 | 92.60 | 70.12 | 70.65 | 69.38 |
| E2E-AFG | 48.48 | 56.92 | 63.24 | 95.45 | 96.14 | 95.67 | 71.47 | 71.80 | 71.62 |

### 4.5 Ablation Studies

Table 3 illustrates the ablation studies conducted on E2E-AFG, assessing the contribution of key components to the overall performance by progressively removing them from the model. First, removing the pseudo-response generation module causes significant performance degradation across multiple tasks, confirming our observation in Sec 3.1: Training data lacking semantic variation features undermines the model's robustness to retrieval noise, particularly manifesting as systematic degradation in evidence evaluation mechanisms when processing semantically deviated contexts. Building on this, further removal of the cross-attention layer in the classification module results in a slight decrease in performance. Without the cross-attention mechanism, the classification module no longer aligns the encoded query $Q$ with the retrieved passages $P$ and pseudo-answers $S$ separately through cross-attention. Instead, $Q$ is concatenated with both representations, and the concatenated features are fed into the feedforward neural network to predict answer existence. Finally, complete removal of the classification module results in substantial performance deterioration, demonstrating that the classification module provides crucial attention guidance to the generator through explicit modeling of contextual credibility distributions, with their synergistic interaction being pivotal to overall performance.

Table 4 validates the differential impact of progressive prompting strategies on the quality of pseudo-answers. The structured Prompt3 performs optimally in logical reasoning tasks, as its structured derivation path provides fine-grained supervisory signals. In contrast, the speculative Prompt2, by allowing reasonable speculation, is better suited for open-domain question answering. Experimental results demonstrate that the noise spectrum constructed by the three prompting strategies exhibits distinct variations in recall rates, with such patterns suggesting potential pathways for enhancing the model's adaptability and robustness in heterogeneous contexts.

Table 5 shows the effect of different top-K retrieved passages on the generation results. Aggregating higher-ranked passages significantly boosts extraction task performance, but at the cost of linearly or quadratically increased computation. However, performance on FEVER and WoW datasets shows no notable improvement, and in some cases declines, likely due to the decreasing quality of lower-ranked passages.

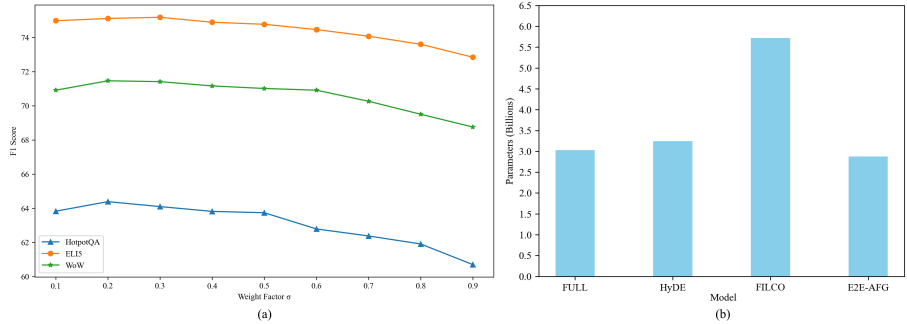

**Fig. 3:** (a) The impact of the weight factor $\sigma$ on model performance. (b) Comparison of model parameters for each method.

Fig. 3(a) illustrates the impact of the weight factor $\sigma$ on model performance. When $\sigma$ is around 0.2 to 0.3, the model achieves optimal performance. As $\sigma$ increases further, the $F_1$ scores across the three datasets begin to decline, with a notable drop when $\sigma$ reaches 0.9. This indicates that in multi-task learning, the distribution of loss weights across different tasks significantly affects model performance, necessitating careful tuning of weight factors for specific tasks.

Fig. 3(b) compares the model parameters for each method. Notably, our proposed E2E-AFG model demonstrates superior parameter efficiency, exhibiting 49.6% fewer parameters compared to FILCO while simultaneously achieving 38.2% reduction in total training duration. This indicates that we maintain model capacity integrity while achieving significant parameter efficiency improvements.

## 5   Conclusion

The End-to-End Model with Adaptive Filtering (E2E-AFG) proposed in this paper effectively addresses the issue of the generator being distracted by irrelevant information retrieved during retrieval-augmented generation tasks. By integrating answer existence judgment with the generation task into a single end-to-end model, E2E-AFG achieves synchronous learning of context filtering and answer generation. Experimental results demonstrate that our model outperforms baseline models across six knowledge-intensive language datasets, with performance improvements ranging from +0.13 to +1.83 points. E2E-AFG not only enhances generation quality but also simplifies model complexity and reduces training costs. Future research could further optimize the model architecture and filtering strategies to explore its potential in various application scenarios.

**Acknowledgments.** This work was supported by the National Key Research and Development Program of China under Grant 2022YFF0903302.